\documentclass{article}

\usepackage[final]{corl_2017} 
\usepackage{bbm}
\usepackage{amsmath}
\usepackage{gensymb}
\usepackage[noend]{algpseudocode}
\usepackage[nothing]{algorithm}  
\usepackage{graphicx}
\usepackage{multirow}
\usepackage{subfigure}

\title{Gradient-free policy architecture search and adaptation}

\author{
	Sayna Ebrahimi\\
	Department of Mechanical Engineering \\
	Department of Electrical Engineering and Computer Sciences\\
	University of California Berkeley, United States\\
	\texttt{sayna@eecs.berkeley.edu} \\
	\And
	Anna Rohrbach \\
	Max Plank Institute for Informatics \\
	Saarland Informatics Campus, Germany \\
	\texttt{arohrbach@mpi-inf.mpg.de} \\
	\AND
	Trevor Darrell \\
	Department of Electrical Engineering and Computer Sciences\\
	University of California Berkeley, United States\\
	\texttt{trevor@eecs.berkeley.edu} \\
}

\begin{document}
	\maketitle
	
	
	\begin{abstract}
		We develop a method for policy architecture search and adaptation via gradient-free optimization which can learn to perform autonomous driving tasks.  By learning from both demonstration and environmental reward we develop a model that can learn with relatively few early catastrophic failures.  We first learn an architecture of appropriate complexity to perceive aspects of world state relevant to the expert demonstration, and then mitigate the effect of domain-shift during deployment by adapting a policy demonstrated in a source domain to rewards obtained in a target environment. We show that our approach allows safer learning than baseline methods,  offering a reduced cumulative crash metric over the agent's lifetime as it learns to drive in a realistic simulated environment.
	\end{abstract}
	\section{Introduction}
	Deep architectures have become popular as function approximators to represent action-selection policies.  Common approaches to learn the parameters of such models include reinforcement learning \citep{sutton1998reinforcement} and/or learning from demonstration \citep{argall2009survey}: both learn model parameters to maximize expected reward, mimic human behavior, and/or achieve implicit goals.  
	However, the design of policy architectures, especially in a deep learning paradigm, remains relatively unexplored. Architectures are typically selected through a combination of intuition and/or trial and error. 
	
	Learning to learn, including the learning of learning architectures, is a long-articulated goal of AI, and many “meta-learning” and “lifelong learning” schemes have been proposed (e.g., \citep{thrun2012learning} offered seminal views; see \citep{schmidhuber2015deep} for a survey). Recently, renewed interest in this topic has focused on models which explicitly search over the structure of deep architectures, including models which fuse non-parametric Bayesian inference with deep learning to select the number of channels for visual recognition tasks \citep{feng2015learning}, models which use reinforcement learning to directly optimize over deep architectures for recognition \citep{zoph2016neural}, and models which use a gradient-free optimization method (“evolutionary search”) to infer optimal network structure \citep{real2017large}.
	
	We investigate policy architecture search using gradient-free optimization and learn optimal policy structure for autonomous driving tasks. We propose a model which learns jointly from demonstration and optimization, with the goal of “safe training”: minimizing the amount of damage a vehicle incurs to learn a threshold level of performance. We base our approach on exploration-based schemes due to their ability to optimize model weights and architecture hyperparameters, leverage expert demonstrations, and adapt to reward obtained in new domains.  We believe that a model which can initialize from demonstration, and learn an optimal policy from that foundation, is likely to achieve higher performance while maintaining the constraint of safe training, compared to models which must randomly search through action space during initial learning, or which learn from a reasonably safe demonstration but cannot further optimize performance based on environmental reward.
	
	Prior approaches to combine demonstration with reward-based learning have had mixed successes \citep{rybski1999interactive,nicolescu2001learning,argall2009survey} mainly due to the poor generalization of the policy learned on demonstrations.  We posit that effective behavior cloning requires learning a visual agent architecture that has sufficient structure to perceive the state of the world deemed relevant to the expert providing the demonstration.  This may or may not be the case with existing, off-the-shelf visual models.  We thus think it is wise to optimize over architectures and parameters when performing expert behavioral cloning. 
	
	Often, deep models which learn to perform in one domain fail to perform well when deployed in another setting, such as differing weather or lighting conditions. 
	Models learned from demonstration are also well known to fail when the learned policy takes the agent away from the region of the state space where the demonstration was provided \citep{ross2011reduction}.
	We show that our method can effectively and safely adapt a model demonstrated in one environment but deployed in a visually different environment based on the reward signal in the latter domain, even when the agent is initialized far from initial demonstrations. Our approach leverages only target domain reward, and makes no assumptions about domain alignment, explicit or implicit, nor assumes any demonstration supervision in the target domain.
	
	To achieve these goals, we present a gradient-free optimization algorithm inspired by \citep{salimans2017evolution} with a modification in noise generation that results in estimating the gradients more efficiently and accurately 
	(Sec. \ref{sec:es}). 
	We then apply this algorithm to search over variable length architectures 
	Next, we combine our gradient-free policy search with demonstrations to learn a better policy that adapts to the new environment by receiving rewards as feedback  (Sec. \ref{sec:task2}). We experimentally show that our architecture search model finds a policy on the GTA game environment that outperforms previously published methods (e.g.,  \citep{bojarski2016end}) in end-to-end  steering prediction from demonstrations, and that it can be efficiently adapted to learn to drive in previously unseen scenarios (Sec. \ref{sec:result}).  Our model reduces the number of crashes incurred while learning to drive, compared to baselines based only on reward or demonstration but not both, or compared to previously proposed fixed architectures that were not optimized for the domain.
	\section{Related work}
	\label{sec:litrev}
	Architecture search has been investigated through different frameworks including reinforcement learning \citep{zoph2016neural,baker2016designing,veniat2017learning} and evolutionary techniques \citep{real2017large}. In \citep{zoph2016neural},  a recurrent neural network (RNN) was used to generate fixed-length architecture descriptions from a predefined search space and trained it with policy gradient methods. They were able to get close and surpass the state of the art results on $\rm{CIFAR}$-$10$ and Penn Treebank datasets, respectively. A meta-modeling algorithm 
	was proposed in \citep{baker2016designing} which used $Q$-learning to sequentially search for convolutional layers for image classification tasks. They showed that their approach outperforms other existing meta-models and manually-designed architectures with similar types of layers. Recently, \citep{veniat2017learning} introduced Budgeted Super Networks which are inspired by the REINFORCE algorithm with an objective function that maximizes prediction quality and computation cost simultaneously.  
	Various versions of biologically-inspired methods, or neuroevolution strategies, have been proposed for architecture search ever since they were introduced by \citep{eigen1973ingo}. Most of them are based on biological genetics algorithms where there is a \textit{fitness} function that gets re-evaluated at each ``generation'' to determine whether ``genotypes'' are perturbed in the correct direction to evolve appropriately  \citep{real2017large,pugh2013evolving,stanley2002evolving,schaffer1992combinations}. I.e., they initialize a model and evolve it based on its performance. This paradigm was recently re-visited as an alternative to reinforcement learning algorithms where optimization is performed in a gradient-free fashion and the algorithm was shown to be highly parallelizable resulting in significant speedups in playing MuJoCo and Atari games \citep{salimans2017evolution}.
	
	Policy search in autonomous driving application has been  largely focused on demonstration-based optimization approaches with \citep{chen2015deepdriving} or without \citep{bojarski2016end,pomerleau1989alvinn} affordance measurements. It dates back to the classic $\rm{ALVIN}$ model \citep{pomerleau1989alvinn} which was a shallow architecture that could map from pixels to simple driving actions. Several years after, researchers demonstrated end-to-end deep learning models for steering control of small-scale cars \cite{lecunoff}, and recently NVIDIA followed the same path and showed success in predicting steering angle on a full-size vehicle from raw pixels using a convolutional network \citep{bojarski2016end}. A novel $\rm{FCN}$-$\rm{LSTM}$ architecture was proposed on large scale crowed-sourced data to perform egomotion predictions conditioned on the previous temporal states \citep{xu2016end}. They used dashcam camera videos to derive a generic driving model that predicted trajectory angle (not steering angles). 
	\section{Approach}
	\label{sec:approach}
	We propose a learning-to-learn model which includes architecture optimization, parameter learning, and representation adaptation over different time scales.
	Our approach can be summarized by the following two steps. (1) Given expert demonstration, search over architectures and parameters to find a policy that best mimics performance by monitoring the obtained accuracy and number of parameters. (2) Having learned from demonstration, adapt the model to the reward provided by the target environment. 
	In both steps, it is essential to derive a function approximator that optimizes an objective function. We use a gradient-free optimization algorithm \citep{salimans2017evolution} that maximizes a parametrized reward function using gradient estimation to perform architecture search (Sec. \ref{sec:task1}) and policy learning (Sec. \ref{sec:task2}). 
	\begin{figure}[t]
		\centering
		\subfigure[]{\label{fig:seen}\includegraphics[height=4.5cm,width=0.5\linewidth]{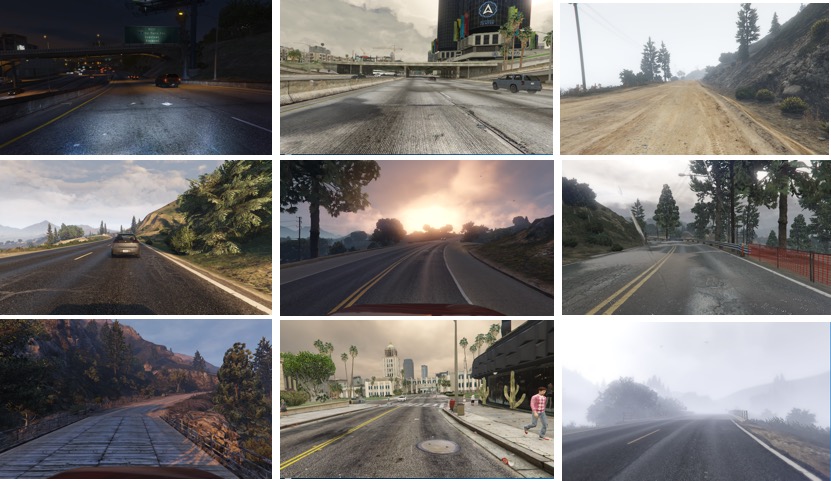}} \quad
		\subfigure[]{\label{fig:unseen}\includegraphics[height=4.5cm,width=0.3\linewidth]{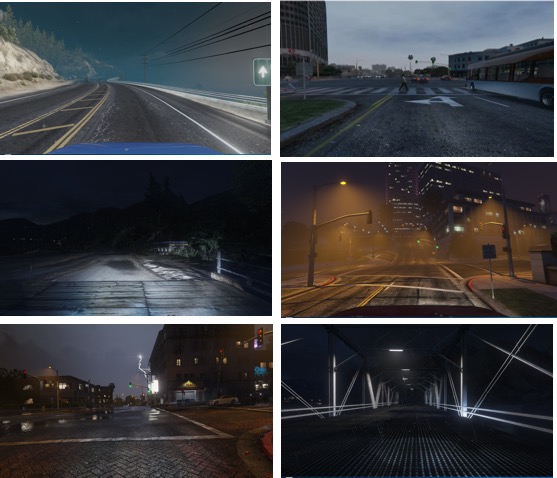}} \\[-3mm]
		\caption{(a) Sample images used in the architecture search for behavioral cloning task; (b) Sample images from a target domain that were not seen during the architecture search for behavioral cloning.}
		\label{fig:samples}
	\end{figure}
	\subsection{Gradient-free optimization algorithm}
	\label{sec:es}
\begin{figure}[t]
		\centering
		\subfigure[]{\label{fig:allnets}\includegraphics[scale=0.3]{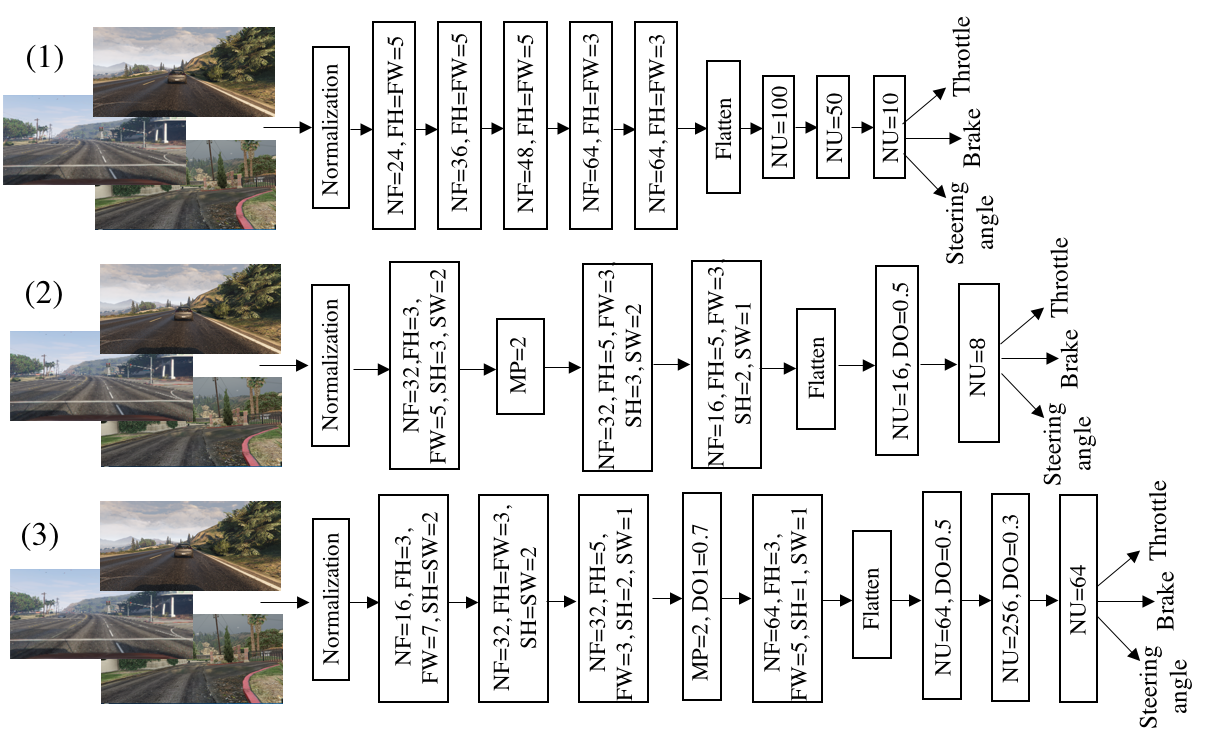}} \quad 
		\subfigure[]{\label{fig:fd-sp}\includegraphics[scale=0.25]{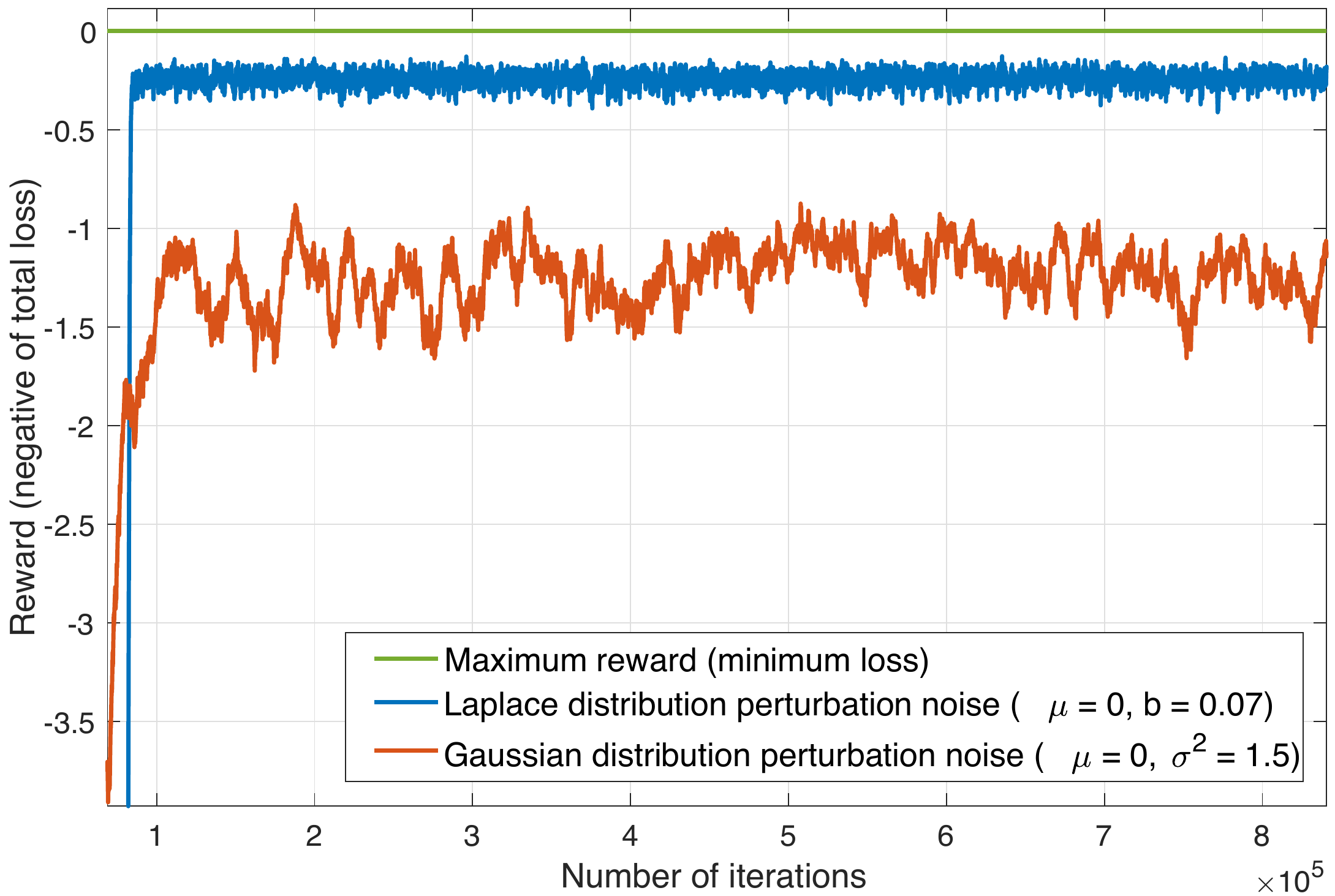}} \\ [-3mm]
		\caption{(a) Illustration of the baseline and learned architectures: (1) prior work \citep{bojarski2016end}, (2) our small network, (3) our large network; (b) Comparison of random Gaussian vs. random Laplace distribution in terms of reward vs. number of iterations.}
		\label{fig:networks}
	\end{figure}

	Let $F(\cdot)$ be our 
	objective function parametrized by $\theta$ which is an $n$-dimensional vector. $F$ can be the reward that an environment provides for an agent when it executes a policy with parameters $\theta$; our goal is to  maximize the expected reward by perturbing the policy parameters, denoted as $\hat{\theta}$, by moving 
	in  particular directions. The parameter estimate update can be performed using a general stochastic form:
	\begin{equation}\label{eq:theta}
	\hat{\theta}_{t+1} = \hat{\theta}_t + \alpha_t ~\nabla_{\hat{\theta}} y(\hat{\theta})
	\end{equation}
	where $y(\cdot)$ is an approximation of the objective function (i.e. $y(\cdot) = F(\cdot) + \rm{noise}$) and $~\nabla_{\hat{\theta}} y(\hat{\theta})$ is the gradient of objective estimate that can be approximated by any gradient estimator in the family of finite difference methods. 
	The gradient is estimated in a randomly chosen direction by perturbing all the elements of $\hat{\theta}_t$ to obtain two measurements of $y(\cdot)$ as follows:
	\begin{align}
	y^{(+)}_t &= F(\hat{\theta}_t + c_t \Delta_t) + \epsilon^{(+)}_t \\
	y^{(-)}_t &= F(\hat{\theta}_t - c_t \Delta_t) + \epsilon^{(-)}_t
	\end{align}
	where $\Delta_t \in R^n$ is a vector of $n$ mutually independent randomly perturbed variables taken from a zero-mean distribution. While there is no restriction for it to have a specific type of distribution, we use Laplace distribution, as it tends to choose orthogonal directions in the long run. Other recent efforts \cite{zoph2016neural} utilized Gaussian noise to sample mirrored projections. Figure \ref{fig:networks} shows a comparison between the two distributions. 
	$c_t$ is a small positive number and $\epsilon^{(+)}_t$ and $\epsilon^{(-)}_t$ are the noise associated with evaluating $F(\cdot)$ such that: $ E \big( \epsilon^{(+)}_t - \epsilon^{(-)}_t | \{\theta_1, \theta_2, \cdots \theta_t \}, \Delta_t \big) = 0 $. The gradient estimate can then be computed as:
	\begin{equation}\label{eq:gradient}
	\nabla_{\hat{\theta_t}} y(\hat{\theta_t}) = \begin{bmatrix}
	\frac{y^{(+)}_t - y^{(-)}_t}{2~c_t~\Delta_{t1}}   & \cdots  & \frac{y^{(+)}_t - y^{(-)}_t}{2~c_t~\Delta_{tn}}           
	\end{bmatrix}^T
	\end{equation}
	Parameter estimates can be updated by replacing the gradients in Eq. \ref{eq:theta} with those found in Eq. \ref{eq:gradient}. 
	\begin{table}[t]
		\centering
		\scriptsize
		\caption{Experimental search space defined for each layer type.}
		\label{tab:space}
		\begin{tabular}{|c|l|}
			\hline
			\multicolumn{1}{|l|}{}                                                                                             & \multicolumn{1}{c|}{Search space}                            \\ \hline
			\multirow{7}{*}{\begin{tabular}[c]{@{}c@{}}Convolutional layer\\ followed by Max-pool and/or Dropout\end{tabular}} & Filter height (FH) $\in$ {[}1, 3, 5, 7{]}                       \\ \cline{2-2} 
			& Filter width (FW) $\in$ {[}1, 3, 5, 7{]}                        \\ \cline{2-2} 
			& Stride height (SH) $\in$ {[}1, 2, 3{]}                         \\ \cline{2-2} 
			& Stride width (SW) $\in$ {[}1, 2, 3{]}                          \\ \cline{2-2} 
			& Number of filters (NF) $\in$  {[}16, 24, 32, 64, 128, 256{]} \\ \cline{2-2} 
			& Max-pool size (MP) $\in$ {[}1, 2, 3{]}                         \\ \cline{2-2} 
			& Dropout (DO1) $\in$ {[}0.3, 0.5, 0.7, 1.0{]}                 \\ \hline
			\multirow{2}{*}{\begin{tabular}[c]{@{}c@{}}Fully-connected layer \\ followed by Dropout\end{tabular}}              & Number of units (NU) $\in$ {[}8, 16, 32, 64, 128, 256, 512{]}    \\ \cline{2-2} 
			& Dropout (DO2) $\in$  {[}0.3, 0.5, 0.7, 1.0{]}                \\ \hline
		\end{tabular}
	\end{table}
	\subsection{Learning an optimal initial policy from demonstrations}\label{sec:task1}
	Inspired by \citep{zoph2016neural} we have used a recurrent neural network to sequentially generate the description of layers of an architecture from a given design space defined by the user. The RNN acts as a controller which generates the architecture description defined by its hyper-parameters chosen from a pre-defined search space. In \citep{zoph2016neural}, the authors used policy gradients to train the RNN which was able to produce fixed-length convolutional and recurrent architectures. Given demonstrations and having a \textit{child} network defined by the RNN, they trained the child network using supervised learning and obtained an accuracy metric on the given task on a held-out validation set and used that accuracy as a reward signal to train the RNN. 
	
	Our model uses the demonstrations to provide the reward function, ((i.e., $F$ above), to train the RNN. Unlike backpropagation which suffers from gradient vanishing while training RNNs, gradient-free algorithms do not have such an issue \citep{salimans2017evolution}. Our RNN controller specifies three types of layers: convolutional, fully connected, and max-pool which can have inter-layer dropouts. For the reward signal, we use the negative value of \textit{total} loss function. At the last layer of the network, we regress to three real-valued numbers, each having a mean-squared loss. The total loss is the sum of all three losses. We use a novel reward function (Eq. \ref{eq:rew} that not only results in the minimum total loss but also grows the architecture as long as the loss keeps decreasing. Note that in case of having a classification problem, an accuracy metric can replace loss value, hence the goal will be maximizing the accuracy while controlling the number of parameters. Here we have a regression problem and our goal is to search for an architecture that is guaranteed to achieve a low loss on the given task and grow further with adding more layers (i.e. producing more parameters) to decrease this while being penalized for adding more parameters in turns of no gain in loss reduction. We propose to use a ReLU-based Lagrange-multiplier reward function as below:
	\begin{equation}
	\label{eq:rew}
	R = R_1 - \lambda\left(R_1\right) R_2
	\end{equation}
	where $R_1$ is the negative of the minimum loss (or maximum accuracy in a classification problem) on the validation set for the last $5$ epochs and $R_2$ is the total number of parameters in the child network. $\lambda$ is the Lagrange parameter defined as a function of the first sub-reward in a ReLU-based fashion:
	\begin{equation} 
	\lambda(R_1) =
	\begin{cases}
	0      & \quad \text{if } R_1 < A \\
	\mu(R_1 - A)  & \quad \text{if } R_1 \geq A \\
	\end{cases}
	\end{equation}
	This reward function acts as follows. The RNN keeps generating new layers by being rewarded only based on the obtained total loss until it produces a child network that achieves the desired value of loss $A$ (or accuracy $A$) on the validation set. Once it reaches this threshold, it will be penalized for further growing the architecture if the loss does not decrease consequently. The parameter $\mu$ in Eq. \ref{eq:rew} defines the respective threshold. E.g. choosing $\mu=0.01/x$ allows architecture growth by the $x$ number of parameters in the new layer if it causes the overall loss decrease (or accuracy increase in classification) by $1\%$. The thresholds are adjustable based on the problem at hand and the desired trade-off between computational cost and loss minimization. 
	
	Our RNN controller is a three-layer LSTM network followed by a softmax layer. The inputs to the RNN are the hyper-parameters that describe a layer (see Tab. \ref{tab:space} for our search space.). Training the RNN starts with randomly initializing the hyper-parameters of the child network which initially has only one layer. RNN uses the reward function to update its parameter weights such that those which contributed more in the obtained reward, receive a higher weighting factor during the update and hence, we move in a hill-climbing direction which eventually maximizes the reward. We use the algorithm described in Sec. \ref{sec:es} to generate an architecture that yields the minimum loss. The process of generating a new layer terminates when we achieve convergence in the received reward. 
	\subsection{Adapting a demonstrated policy to a new driving domain}
	\label{sec:task2}
	As confirmed experimentally below, it is well known that a policy learned from behavioral cloning can perform poorly when evaluated on inputs with a domain shift relative to the demonstration supervision.  To overcome this, we further use the gradient-free search algorithm described in Sec. \ref{sec:es} to adapt a driving policy learned from demonstration in a source domain based on rewards in a target domain. We experiment with the setting where the initial agent state (e.g., location), and/or weather and lighting conditions, are substantially different than provided as demonstration. 
	We compare to baselines where we perform reward-based optimization using initial demonstrations instead of a randomly initialized policy, which makes our reward function converge faster, and more safely. 
	
	In the driving scenario, we wish to learn to drive with optimal or near-optimal performance, defined by the reward in the target domain. Specifically, the reward function used in our experiment is composed of two factors (we receive $+1$ if obeyed and $0$ if violated): 1) No crashes with other objects 2) Staying within the lane lines if they are available in the driving scene. We have used the lane reward function and accident detection function defined in \citep{gtaplugin} source code. The necessary information 
	is provided by a $\textit{paths.xml}$ file in \citep{gtaplugin}.
	
	In our model, an \textit{episode} is the time interval that the agent has successfully driven without having a car crash. Note that not all deviations from the middle of the road necessarily result in an accident. In case of a minor deviation, while the car receives $0$ as its reward, it continues driving until it makes a mistake that causes it to crash and the game restarts. There are distinct thresholds for middle-lane deviation defined in \citep{gtaplugin} for different roads (highway, urban, etc.) and different vehicle types.
	\section{Experimental evaluation}\label{sec:result}
	We implemented the method described above and ran comprehensive experiments to show the efficiency and applicability of our approach in searching for an optimal driving policy that has the minimum number of catastrophic failures. Details of the experiments along with the results are provided in the following subsections. All the experiments are executed in the GTA game environment using a publicly available plugin \citep{gtaplugin} that allowed us to have control over driving conditions such as lighting, weather, car model, and reward function.
	\vspace{-10pt}
	\subsection{Dataset} For our evaluation we have collected a dataset of an expert policy by playing GTA collecting $2,267,662$ images of size $66 \times 200$ similar to \citep{bojarski2016end}. Labels include steering angle, brake, and throttle values. 
	In order to learn from diverse driving scenarios, we have used data from different locations (highways, rural roads, urban streets) where weather and lighting conditions were adjusted using \citep{gtaplugin}. Our goal was to expose the learning algorithm to a comprehensive demonstration set yet to set aside some specific scenes for further testing the performance of behavioral cloning task. Sample images from demonstration are shown in Fig. \ref{fig:seen}. They include rainy (daytime), overcast (day and night), foggy (daytime), sunny scenes and thunderstorms (daytime). Some particular scenes such as rain and thunderstorm during nighttime as well as snow at daytime have been kept for our test set (see Fig. \ref{fig:unseen}). Our test set is composed of $100,000$ images. 
	
	\subsection{Learning a policy architecture from GTA demonstrations} 
	The search space for the hyperparameters that describe a fully-convolutional architecture is presented in Tab. \ref{tab:space}. The activation function is fixed to be a rectified linear unit. 
	The RNN-controller has three LSTM layers, with $10$ hidden units in each, and a softmax layer at the end to choose from the given search space. The RNN weights are initialized with a random Laplace distribution $(\mu=0,b=0.07)$. Once the RNN predicts a new layer's description, the child network is built and trained with batch size of $64$ and Adam optimizer \citep{kingma2014adam} with learning rate $10^{-3}$. We train the child network for different number of epochs starting from $20$ epochs (depending on which layer we are at) and compute the reward function as described in Sec. \ref{sec:task1}.  In order to finish optimizing one layer, we track the loss reduction of both validation and training sets between the first and the last epoch to avoid overfitting. 
	
	Our model is capable of generating architectures at low or high costs of architecture growth. In order to compare our designed architecture with \citep{bojarski2016end},\footnote{As no reference implementation of  \citep{bojarski2016end} is openly available we had to use our own, which may be suboptimal w.r.t. the authors' as we did not have full access to their model parameters. Also, their model was only used to predict steering angle, and overall it is not clear whether their goal was to maximize  performance, find a model  with relatively few parameters, or both, so they may not have explored the full design space with their model. Nonetheless, \citep{bojarski2016end} was the closest model in the literature for end-to-end steering angle prediction and thus the best available baseline.} we control the reward function such that it never produces an architecture with more than $252,241$ parameters while minimizing the loss (maximizing the reward). We present our architectures and comparison to prior work in Fig. \ref{fig:allnets}(2). The corresponding performance comparison is shown in Tab. \ref{tab:bc_results}. The smallest architecture (in terms of the number of parameters) that we have built is shown in Fig. \ref{fig:allnets}, which obtains a smaller total loss than \citep{bojarski2016end}. The network of \citep{bojarski2016end} (Fig. \ref{fig:allnets}(1)) appears to suffer from overfitting which might be explained by the absence of Dropout or a pooling layer. Not restricting our architecture search algorithm to be bounded by a number of parameters, we learn a larger network (\ref{fig:allnets}(3)) that has over $2M$  parameters and obtains a minimum total loss of $0.085$ on the training set and $0.088$ on the validation set. As discussed above, we test all models on substantially different driving scenes which were never seen during training. On this challenging test set our large network obtains the minimum loss of $0.195$. We use this network as our initial driving policy in the next section and improve it further in the reward-providing GTA environment. 
	\begin{table}[h]
		\centering
		\footnotesize
		\caption{Total MSE obtained using architecture proposed in prior work \citep{bojarski2016end} and our models obtained by architecture search on demonstrations.}
		\label{tab:bc_results}
		\begin{tabular}{|l|c|c|c|c|c|}
			\hline
			Model                   &  $\#$ of parameters &  Training loss & Validation loss &  Test loss \\ \hline 
			Bojarski et al. \citep{bojarski2016end} & 252,241 & 0.098 & 0.11 & 0.212 \\ \hline
			Our small network & 228,227 & 0.093 & 0.096 &  0.197 \\ \hline
			Our large network &  2,198,723 & 0.085 & 0.088 & 0.185 \\ \hline
		\end{tabular}
	\end{table}
	
	We further empirically compare two different noise distributions for perturbing the parameters of the network: random Gaussian and random Laplace. We perform the architecture search over our small model using both distributions with mean zero; variance is chosen using grid search. Fig. \ref{fig:fd-sp} shows the results for reward convergence versus the number of iterations. Both distributions result in convergence to high reward values (minimizing loss), however, the Laplace distribution tends to be less noisy and reaches slightly higher reward values.
	
	\subsection{Safe policy adaptation}
	Next, we want to learn the driving policy in a target game domain. We start with an initial model, either using a behaviorally cloned or randomly initialized policy and gradually improve it by receiving rewards from the environment.  As stated in \ref{sec:task2}, an episode of the game starts at a random location and weather condition in the game. To initialize the policy, we use the larger architecture learned in the first step (with the model of \citep{bojarski2016end} as the baseline). We evaluate both models with and without being adapted to demonstration forming four cases: (1) the baseline network of without demonstration (i.e., with randomly initialized weights) and (2) with behaviorally cloned initial weights, (3) our larger architecture without demonstration and (4) with behaviorally cloned initial weights. We run all models in the GTA environment to receive the reward described in Sec. \ref{sec:task2}. Once the reward is received, the weights are perturbed by a Laplace random noise $(\mu=0,b=0.07)$ and the same procedure is repeated until the average reward in each episode of the game converges to its maximum value. Results averaged across several runs are presented in Tab. \ref{tab:task2} where our model optimized with demonstration outperforms all other cases. In particular, our model has the least number of cumulative crash occurrence prior to converging to $100\%$ of averaged reward (details below). 
	\begin{table}[t]
		\footnotesize
		\centering
		\caption{Comparison of two policies (our large network and \citep{bojarski2016end}) learned based on target domain reward, with and without source-domain demonstrations.}
		\label{tab:task2}
		\begin{tabular}{|c|c|c|c|}
			\hline
			Model  & \begin{tabular}[c]{@{}c@{}}Wall-clock \\ convergence \end{tabular} & \begin{tabular}[c]{@{}c@{}}Total \# \\ of car crashes \end{tabular} & \begin{tabular}[c]{@{}c@{}}Total \# of middle-lane \\ keeping violations\end{tabular} \\ \hline
			\multicolumn{1}{|l|}{(1) Bojarski et al. \citep{bojarski2016end}, w.o demo}& 154 hours                                                                                      & 15,565                     & 18,662                                                                                                            \\ \hline
			\multicolumn{1}{|l|}{(2) Bojarski et al. \citep{bojarski2016end}, w. demo}& 74 hours                                                                                      & 1,387                      & 3,243                                                                                                            \\ \hline
			\multicolumn{1}{|l|}{(3) Our large network, w.o. demo} & 114 hours                                                                                      & 6,877                      & 8781                                                                                                             \\ \hline
			\multicolumn{1}{|l|}{(4) Our large network, w. demo}& 53 hours                                                                                      & 832                       & 982                                                                                                              \\ \hline
		\end{tabular}
	\end{table}

	In Table \ref{tab:adaptation} we have listed the results for test accuracies on a dataset taken from a target domain that has not been seen in the demonstrations.  On the left, we see behavioral cloning alone has poor performance when significant domain shifts occur. Our adapted model has performance over loss minimization of $10^{-5}$. On the right we see that adapted performance is strong even without reward in the target domain, indicating that visual domain shift is a lesser issue than being off-demonstration; our model can adapt in the source domain and still be accurate on the target.  Best performance is obtained with adaptation to reward in both source and target. This also shows that there is an improvement on loss minimization when we learn from rewards. It is worth noting that we can not judge the driving behavior only by looking at the total MSE loss as it is not a comprehensive representative of the driving task. Each one of the angle, brake, and throttle converges to a separate MSE loss among which steering angle has the least and brake has the largest loss values. This shows that learning steering angle is easier with demonstrations compared to brake and throttle which at each time step depend on multi previous frames.
	Table \ref{tab:ped} shows all model predictions for a relatively complex image chosen from the target domain (Fig. \ref{fig:samples} at top corner) where a pedestrian crossing the street when the signal light is green. The behaviorally cloned models tend to predict that the agent should keep going whereas the adapted models to the target domain with rewards are able to predict the correct decision despite the green light presence in the image. Our adapted large network rewarded on both source and target is able to make the best prediction for throttle and brake (steering angle is perfect across all models).  
	\begin{table}[t]
		\centering
		\scriptsize
		\caption{Comparison of performance (average total loss) of two policies (our large network and \citep{bojarski2016end}) at test time on target domain (T) when they are trained with rewards from target domain, source domain (S), and both (T+S).}
		\label{tab:adaptation}
		\begin{tabular}{l|c|c|llcccc}
			\cline{2-3}
			& \begin{tabular}[c]{@{}c@{}}Behavioral \\ cloning\end{tabular} & \begin{tabular}[c]{@{}c@{}}Demo on S\\ Rew. on T+S\\ Test on T\end{tabular} &                       &                                                                                   & \multicolumn{1}{l}{}                                                               &                                                                                                & \multicolumn{1}{l}{}                                                                             & \multicolumn{1}{l}{}                                                                           \\ \cline{1-3} \cline{6-9} 
			\multicolumn{1}{|l|}{Bojarski et. al.\citep{bojarski2016end}}                                    & 0.212                                                       & $10^{-4}$                                                                     &                       & \multicolumn{1}{c|}{}                                                             & \multicolumn{1}{c|}{\begin{tabular}[c]{@{}c@{}}Behavioral \\ cloning\end{tabular}} & \multicolumn{1}{c|}{\begin{tabular}[c]{@{}c@{}}Demo on S\\ Rew. on S\\ Test on T\end{tabular}} & \multicolumn{1}{c|}{\begin{tabular}[c]{@{}c@{}}Demo on S\\ Rew. on S+T\\ Test on T\end{tabular}} & \multicolumn{1}{c|}{\begin{tabular}[c]{@{}c@{}}Demo on S\\ Rew. on T\\ Test on T\end{tabular}} \\ \cline{1-3} \cline{5-9} 
			\multicolumn{1}{|l|}{\begin{tabular}[c]{@{}l@{}}Our large \\ network\end{tabular}} & 0.185                                                       & $10^{-5}$                                                                     & \multicolumn{1}{l|}{} & \multicolumn{1}{l|}{\begin{tabular}[c]{@{}l@{}}Our large \\ network\end{tabular}} & \multicolumn{1}{c|}{0.185}                                                       & \multicolumn{1}{c|}{$7\times10^{-5}$}                                                                   & \multicolumn{1}{c|}{$10^{-5}$}                                                                       & \multicolumn{1}{c|}{$3\times10^{-5}$}                                                                   \\ \cline{1-3} \cline{5-9} 
		\end{tabular}
	\end{table}
	
\begin{figure}[t]
		\centering
		\subfigure[]{\label{fig:trained}\includegraphics[scale=0.34]{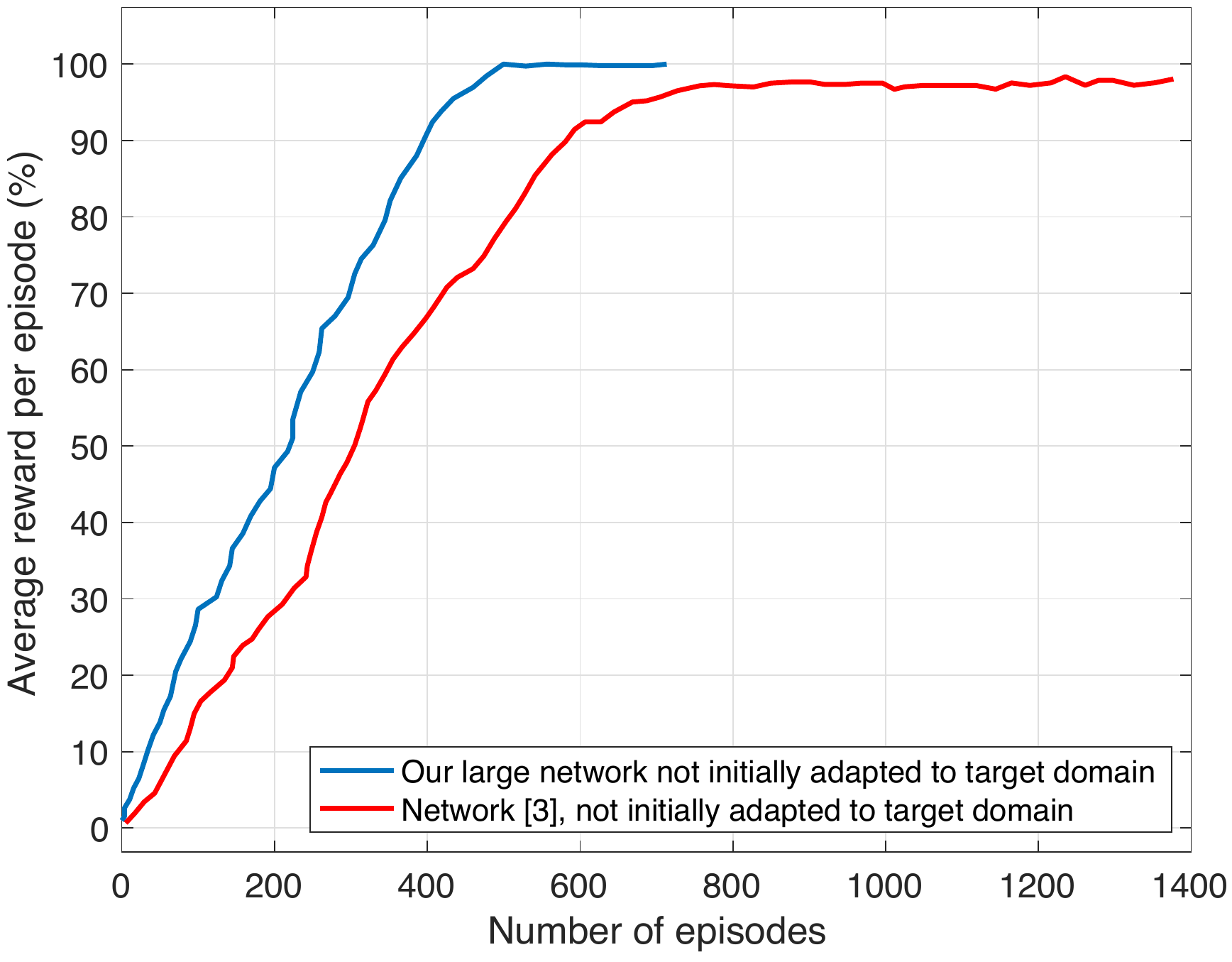}} \quad
		\subfigure[]{\label{fig:randoms}\includegraphics[scale=0.34]{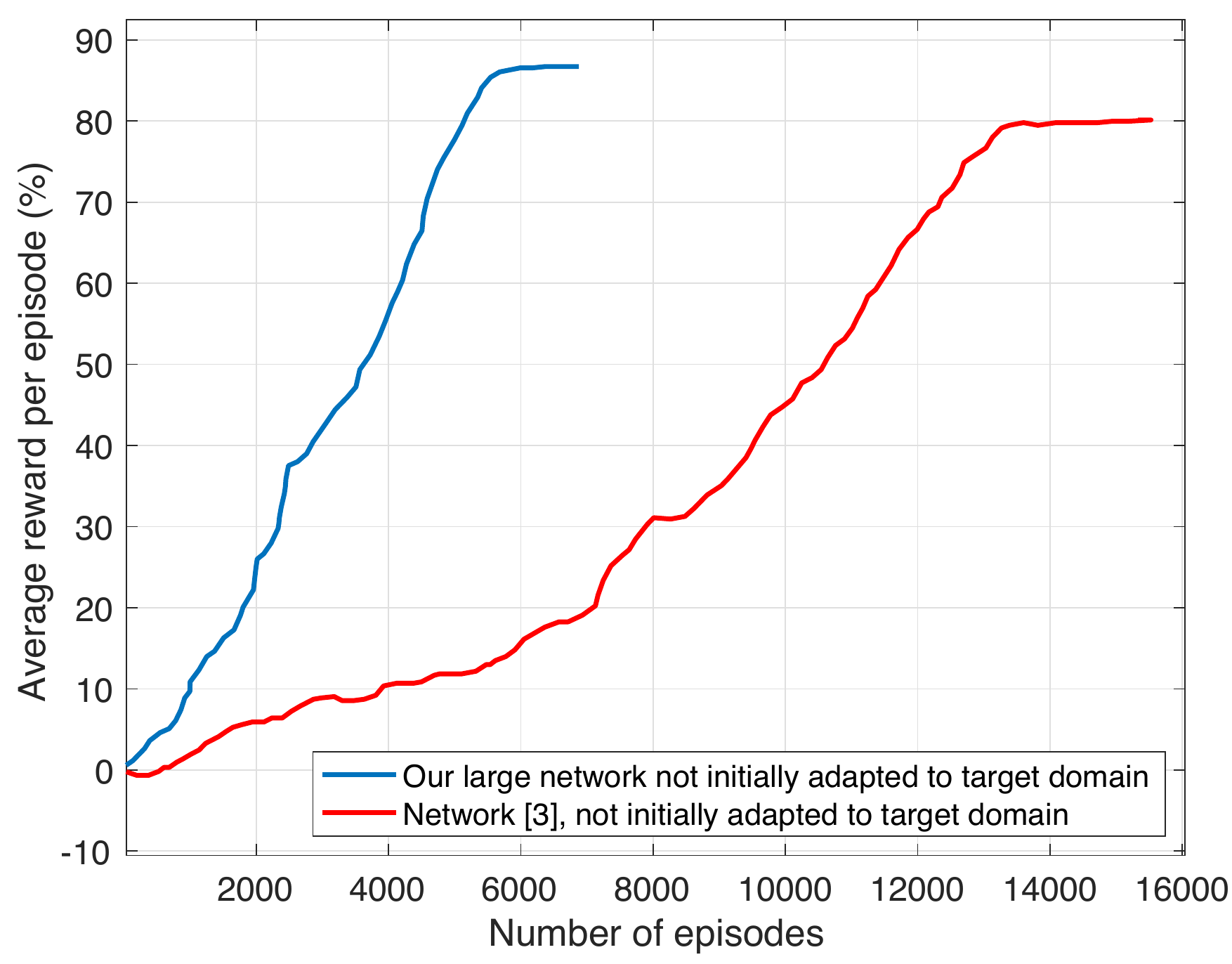}} \\ [-4mm]
		\caption{Averaged reward per episode vs. number of episodes for our large network and \citep{bojarski2016end} using (a) policy adapted to demonstrations; (b) a non-adapted policy. (Note different x-axis scales.)}
		\label{fig:res_task2}
	\end{figure}

	Fig. \ref{fig:res_task2} illustrates the percentage of averaged reward per episode for the aforementioned four models until convergence. Our designed architecture which is adapted to the demonstrations on a source domain, starts with less than $10\%$ reward in its first episode which lasts for $37$ seconds. This is reasonable considering the fact that each episode of the game, is intentionally set up to start in a completely random new environment which is highly possible to be a significantly different domain that what the policy has seen up to that point. This again highlights the fact that a behaviorally cloned model is at high risks of failure when it is tested in a different domain. The models keep learning from the rewards until convergence. It can be seen in Fig. \ref{fig:trained} that our designed architecture adapted to the demonstration reaches $100\%$ of averaged reward after $53$ hours in its last episode (episode number $\#832$) which lasts for $90$ minutes and is then terminated by the user (no crashing happens). It is also shown that a suboptimal, yet adapted to the demonstrations policy \citep{bojarski2016end}, also converges but only to $97.3\%$ of the maximum reward and then plateaus for more than $12$ hours. Unadapted policies are also shown in Fig. \ref{fig:randoms} converging to an averaged reward of $87\%$ in a drastically different time-scale  confirming the positive effect of using demonstrations in policy learning.
	Supplementary video of results can be found in \url{https://saynaebrahimi.github.io/corl.html}
	\begin{table}[H]
		\label{tab:ped}
		\scriptsize
		\centering
		\caption{Predictions for a complex driving scene shown in Figure \ref{fig:samples} (top right corner) using all our learned models} 
		\label{tab:ped}
		\begin{tabular}{cl|c|c|c|c|}
			\cline{3-6}
			&                & \begin{tabular}[c]{@{}c@{}}Behavioral\\ cloning\end{tabular} & \begin{tabular}[c]{@{}c@{}}Demo on S\\ Rew on S\\ Test on T\end{tabular} & \begin{tabular}[c]{@{}c@{}}Demo on S\\ Rew on S+T\\ Test on T\end{tabular} & \begin{tabular}[c]{@{}c@{}}Demo on S\\ Rew on T\\ Test on T\end{tabular} \\ \hline
			\multicolumn{1}{|c|}{\multirow{3}{*}{\begin{tabular}[c]{@{}c@{}}Large\\ network\end{tabular}}} & Steering angle & -0.006                                                       & 0.003                                                                    & 0.005                                                                      & 0.002                                                                    \\ \cline{2-6} 
			\multicolumn{1}{|c|}{}                                                                         & Brake          & 0.191                                                        & 0.889                                                                    & 0.931                                                                      & 0.956                                                                    \\ \cline{2-6} 
			\multicolumn{1}{|c|}{}                                                                         & Throttle       & 0.665                                                        & 0.083                                                                    & 0.010                                                                      & 0.052                                                                    \\ \hline
			\multicolumn{1}{|c|}{\multirow{3}{*}{Bojarski et. al. {[}3{]}}}                                & Steering angle & -0.005                                                       & -0.002                                                                   & 0.002                                                                      & -0.001                                                                   \\ \cline{2-6} 
			\multicolumn{1}{|c|}{}                                                                         & Brake          & 0.183                                                        & 0.567                                                                    & 0.677                                                                      & 0.778                                                                    \\ \cline{2-6} 
			\multicolumn{1}{|c|}{}                                                                         & Throttle       & 0.775                                                        & 0.223                                                                    & 0.121                                                                      & 0.156                                                                    \\ \hline
		\end{tabular}
	\end{table}
	\section{Conclusion}
	\label{sec:conclusion}
	The goal of this work is to learn an  policy for an autonomous driving task  minimizing crashes and other safety violations while training. To this end we propose an  algorithm which  learns to generate an optimal network architecture from demonstration using a new reward function that optimizes accuracy and model size simultaneously. We confirm behavioral cloning alone can perform poorly when the target domain differs from source demonstrations. We show that our method can  adapt the model learned by demonstration to a new domain relying on  target environmental rewards. Experimental evaluation shows that  our model achieves higher accuracy, fewer cumulative crashes, and higher target domain reward. We believe these results are encouraging and important steps towards the  ultimate goal of learning complex driving policies with zero cumulative crashes or serious accidents either in simulation or the real world.
	
	
	
	\clearpage
	

	\bibliography{example}  
	
\end{document}